# The Minimum Information Principle for Discriminative Learning

**Amir Globerson**　　　　**Naftali Tishby**

School of Computer Science and Engineering and
Interdisciplinary Center for Neural Computation
The Hebrew University, Jerusalem 91904, Israel

## Abstract

Exponential models of distributions are widely used in machine learning for classification and modelling. It is well known that they can be interpreted as maximum entropy models under empirical expectation constraints. In this work, we argue that for classification tasks, mutual information is a more suitable information theoretic measure to be optimized. We show how the principle of minimum mutual information generalizes that of maximum entropy, and provides a comprehensive framework for building discriminative classifiers. A game theoretic interpretation of our approach is then given, and several generalization bounds provided. We present iterative algorithms for solving the minimum information problem and its convex dual, and demonstrate their performance on various classification tasks. The results show that minimum information classifiers outperform the corresponding maximum entropy models.

## 1 INTRODUCTION

The use of probabilistic methods in classification - predicting the class $y$ for the variable $x$ - is widespread. Such methods form a model of $p(y|x)$ and use it to predict $y$ from $x$. It is customary to divide such methods into two classes: Generative and Discriminative. Generative methods approximate the joint distribution $p(x, y)$ and use Bayes rule to obtain $p(y|x)$. Often $p(y)$ is assumed to be known and the class conditional distributions $p(x|y)$ are estimated separately. On the other hand Discriminative models approximate $p(y|x)$ directly. The latter have the clear advantage of solving the classification problem directly.

One of the advantages of generative models, however, is that the only data they use are within class statistics, such as the class mean and variance in Gaussian models, or conditional marginals in Naive Bayes models. This property endows them with attractive convergence properties, since such statistics often demonstrate fast convergence to their true values [15]. On the other hand, the fact that $p(x|y)$ is formed independently for each value of $y$ prevents these models from finding features which specifically discriminate between the two classes. In this work we present a modeling principle which uses only class conditional statistics, but does consider their interactions for different classes.

One of the most common forms of class conditional models in the literature is the exponential form

$$p(x|y) = \frac{1}{Z_y} e^{\sum_i \psi_i(y)\phi_i(x)} , \qquad (1)$$

where the functions $\phi_i(x)$ are given, and $\psi_i(y)$ are determined via maximum likelihood estimation. One motivation often cited for using exponential distributions is their link to the Maximum Entropy (MaxEnt) principle [11]. The distribution in Equation 1 maximizes the entropy over the set of distributions which share the expected values of $\vec{\phi}(x)$.

However, when considering the task of classification, or regression, one wishes to use an optimization criterion which is directly related to the prediction of $Y$ from $X$, rather than to the distribution $p(x, y)$. A fundamental quantifier of the prediction quality is the mutual information $I(X;Y)$ [2], which is a model free quantifier of the dependence between $X$ and $Y$. Furthermore, it provides a bound on the optimal Bayes error [8], among its other important properties.

In distributional inference, MaxEnt searches for the "least committed" distribution agreeing with the empirical constraints. The equivalent concept in classification would be the distribution under which $X$ provides the *least* information about $Y$. This suggests looking for the distribution with minimal mutual information. In what follows, we denote this principle by MinMI.

When the prior distributions $p(x)$ and $p(y)$ are known, MinMI is equivalent to MaxEnt for the joint $p(x,y)$. However, when one of the priors is unknown, MinMI corresponds to a different principle which, unlike maximum entropy, is *not* equivalent to maximum likelihood for any parametric model.

This principle is also related to the Information Bottleneck method [17], where information minimization is used as in Rate-Distortion theory to obtain efficient data representations, and to Sufficient Dimensionality Reduction (SDR) algorithm [6], where information minimization is used to derive a feature extraction algorithm. However, in SDR both marginals are assumed to be known. This poses a difficulty for large $X$ spaces, which is solved in the current work. Another work which connects learning and maximum entropy is the Maximum Entropy Discrimination (MED) framework [10], where a maximum entropy prior over the space of classifiers is sought. Minimization of Mutual Information is also at the basis of the Independent Component Analysis (ICA) method, which is not related to classification.

## 2 Problem Formulation

We now define the minimum information problem, and characterize its solution. Let $y_i$ be the class of the sample $x_i$ for $i = 1 \ldots n$. The empirical class marginal is

$$\bar{p}(y) \equiv \frac{1}{n} \sum_i \delta_{y_i, y} . \qquad (2)$$

Let $\vec{\phi}(x) : X \to \Re^d$ be a given function of $X$. The class conditional empirical means of $\vec{\phi}(x)$ are then

$$\vec{a}(y) \equiv \frac{1}{n\bar{p}(y)} \sum_{i: y_i = y} \vec{\phi}(x_i) . \qquad (3)$$

We now calculate the distribution which has minimum mutual information while agreeing with the sample on both the expected values of $\vec{\phi}(x)$ for each class, and the marginal $p(y)$. Define the set of distributions agreeing with the sample by

$$\mathcal{P}(\vec{a}) \equiv \left\{ p(x,y) : \begin{array}{l} \langle \vec{\phi}(x) \rangle_{p(x|y)} = \vec{a}(y) \quad \forall y \\ p(y) = \bar{p}(y) \end{array} \right\} . \qquad (4)$$

The information minimizing distribution is then given by

$$p_{MI}(x,y) \equiv \arg \min_{p(x,y) \in \mathcal{P}(\vec{a})} I[p(x,y)] , \qquad (5)$$

where $I[p(x,y)]$ denotes the Mutual Information between $X$ and $Y$ under the distribution $p(x,y)$. Note that since the marginal $p(y)$ is constrained, we are actually optimizing over $p_{MI}(x|y)$. This minimization problem is convex since the Mutual Information is a convex function of $p(x|y)$ for a fixed $p(y)$ [2] and the set of constraints is also convex. It thus has no local minima.

Using Lagrange multipliers to solve the constrained optimization in Equation 5 we obtain the following characterization of the solution,

$$p_{MI}(x|y) = p_{MI}(x) e^{\vec{\psi}(y) \cdot \vec{\phi}(x) + \gamma(y)} , \qquad (6)$$

where $\vec{\psi}(y)$ are the Lagrange multipliers corresponding to the constraints, and $\gamma(y)$ is set to normalize the distribution. Note that this does not provide an analytic characterization of $p_{MI}(x|y)$ since $p_{MI}(x)$ itself depends on $p_{MI}(x|y)$ through the marginalization

$$p_{MI}(x) = \sum_y p_{MI}(x|y) \bar{p}(y) . \qquad (7)$$

The minimum mutual information has the following simple expression

$$I[p_{MI}(x,y)] = \langle \gamma(y) + \vec{\psi}(y) \cdot \vec{a}(y) \rangle_{\bar{p}(y)} \qquad (8)$$

where the operator $\langle \rangle_{\bar{p}(y)}$ denotes expectation with respect to $\bar{p}(y)$.

In performing prediction of the class variable $Y$, we will be using the distribution $p_{MI}(y|x)$ as a *plug-in* estimate of $p(y|x)$. By Bayes law we have [1]

$$p_{MI}(y|x) = \bar{p}(y) e^{\vec{\psi}(y) \cdot \vec{\phi}(x) + \gamma(y)} . \qquad (9)$$

Note that this distribution has a form similar to logistic regression as in [13]. However, there are two main differences between $p_{MI}(y|x)$ and the standard logistic regression. One is that $p_{MI}(y|x)$ does not have a normalization function dependent on $X$. This is a common property of information minimizing distributions, and is also seen in Rate Distortion theory [2] and the Information Bottleneck method [17]. The second difference is that the optimal parameters of $p_{MI}(y|x)$ are not those obtained via (conditional) maximum likelihood, but rather those which satisfy the conditions in Equation 6. This constitutes another difference between our formalism and that of MaxEnt, which is known to be equivalent to Maximum Likelihood estimation in exponential models [5].

### 2.1 A Dual Problem

The constrained information minimization in Equation 5 is a feasible convex optimization problem, and therefore has an equivalent Lagrange dual. The dual for a similar problem - finding the Rate Distortion function , was recently shown to be a geometric program [1].

---
[1] Note that dividing by $p_{MI}(x)$ is allowed only if it is non-zero. If $p_{MI}(x) = 0$ the function in Equation 9 may not be normalized. However, we may still use the resulting unnormalized $p_{MI}(y|x)$ to perform classification.

Using similar duality transformations to those in [1], we obtain the following geometric program (in convex form), which is equivalent to the MinMI problem in Equation 5,

$$\begin{aligned}\text{Maximize} \quad & \langle \gamma(y) + \vec{\psi}(y) \cdot \vec{a}(y) \rangle_{\bar{p}(y)} \\ \text{Subject To} \quad & \log \sum_y \bar{p}(y) e^{\gamma(y) + \vec{\psi}(y) \cdot \vec{\phi}(x)} \leq 0 \quad \forall x\end{aligned}.$$
(10)

Optimization is over the variables $\gamma(y), \vec{\psi}(y)$, and there are $|X|$ constraints. The maximum of Equation 10 is equal to the minimum information obtained in Equation 5. Another interesting property of the dual problem is that the inequality constraints are not strict only for $x$ such that $p_{MI}(x|y) = 0$ in the primal problem. This is a direct result of the Kuhn-Tucker conditions.

In section 7 we discuss algorithmic solutions to both the primal and the dual problems.

## 3 A Game Theoretic Interpretation

In [7] Grunwlad gives a game theoretic interpretation of the MaxEnt principle. We now describe a similar interpretation which applies to the MinMI principle. The following result can be proven using arguments similar to those in [7] [2] .

**Proposition 1** *Let $\mathcal{A}$ be the set of all distributions of $Y$ conditioned on $X$. If $p_{MI}(x) > 0$ for all $x$, the minimum information distribution satisfies*

$$p_{MI}(y|x) = \arg \min_{q(y|x) \in \mathcal{A}} \max_{p(x,y) \in \mathcal{P}(\vec{a})} -\langle \log q(y|x) \rangle_{p(x,y)} .$$

The above proposition implies that $p_{MI}(y|x)$ is obtained by playing the following game: Nature chooses a distribution $p(x,y)$ from $\mathcal{P}(\vec{a})$. The player, who does not know $p(x,y)$ then chooses a conditional distribution $q(y|x)$ aimed at predicting $Y$ from $X$. The loss incurred in choosing $q(y|x)$ is given by $-\langle \log q(y|x) \rangle_{p(x,y)}$. The proposition states that $p_{MI}(y|x)$ corresponds to the strategy which minimizes the worst case loss incurred in this game.

To see how the above argument is related to classification error, we focus on the binary class case, and take the class variable to be $y = \pm 1$. In this case, a classifier based on $q(y|x)$ will decide $y = 1$ if $q(y=1|x) \geq 0.5$. The zero-one loss is thus

$$c_{zo}(x,y,q) = \Theta\big[-\big(q(y=1|x) - 0.5\big)y\big] , \quad (11)$$

---
[2]The result assumes strict positivity of $p_{MI}(x)$. We are currently investigating whether this condition can be relaxed.

where $\Theta$ is the step function , and $y$ is the *true* label for $x$. The zero-one loss is bounded from above by the loss function $-\log_2 q(y|x)$

$$c_{zo}(x,y,q) \leq -\log_2 q(y|x) . \quad (12)$$

The classification error incurred by $q$ is thus bounded from above by the expected loss

$$\langle c_{zo} \rangle_{p(x,y)} \leq \langle -\log_2 q(y|x) \rangle_{p(x,y)} \quad (13)$$

Note that for the information minimizing distribution $p_{MI}(y|x)$ the above loss is the familiar logistic loss.

We thus have the following elegant formulation of MinMI: the *plug in* distribution $p_{MI}(y|x)$ is the one which minimizes the worst case upper bound on classification error.

## 4 MinMI and Joint Typicality

The rationale for the MaxEnt principle, as given by Boltzmann, Jayens and others, is based on the fact that samples with atypical empirical histograms - hence with lower empirical entropy - are exponentially (in the sample size) unlikely to occur. Thus we can assert by a histogram counting argument that out of all histograms consistent with observed expectation values, those with maximum entropy are the most likely to be observed among all consistent histograms in the absence of any other knowledge.

When dealing with classification or regression problems, the issue is predictions of $Y$ from $X$, and it is the notion of *joint typicality* of the two sequences that replaces the simple typicality and AEP property in the MaxEnt case. Here we are asking for the most uncommitted distribution of $x$, *given* that we know the margin distribution of $y$, $p(y)$, together with a set of empirical conditional expectations. For this case a similar histogram counting argument is supplied through the notion of joint typicality, as stated e.g. in [2] pp. 359.

Let $Y^n = Y_1, Y_2, ..., Y_n$ be drawn i.i.d. from $\prod p(y)$. Then for any sequence $x^n = x_1, x_2, ..., x_n$, the probability that $(x^n, Y^n)$ are jointly drawn i.i.d. from $p(x,y)$ is $\simeq 2^{-nI(X;Y)}$, via the standard AEP property. In other words, if we partition all the possible empirical histograms of $x^n$ into equivalent classes according their (empirical) mutual information with $Y^n$, $I(X;Y)$, the relative volume of such a class is exponential in its mutual information and proportional to $2^{-nI(X;Y)}$.

Without any other constraints the (overwhelmingly) largest joint-histogram of $x^n$ and $Y^n$ is the one with $I(X;Y) = 0$, i.e. independent $X$ and $Y$. Otherwise, with additional empirical constraints on the joint distribution, the overwhelming large fraction among the $x^n$ histograms is occupied by the one with the minimal empirical mutual information. This is the distribution selected by our proposed MinMI procedure.

## 5 Generalization Bounds

The Minimum Information principle suggests a parsimonious description of the data, and therefor one would expect it to have generalization capabilities. We discuss several generalization related results below. To simplify the discussion, we focus on the binary class case. Denote by $p(x,y)$ the *true* distribution underlying the data. Also, denote by $e^*(p)$ the optimal Bayes error associated with $p(x,y)$, and $e_{MI}$ the generalization error when using $p_{MI}(y|x)$ for classification.

The Bayes error $e^*(p)$ is the minimum classification error one could hope for, when predicting $y$ from $x$ under $p(x,y)$. The following Lemma [8] bounds the Bayes error using the Mutual Information

$$e^*(p) \leq \frac{1}{2}\left(H(Y) - I[p(x,y)]\right) . \qquad (14)$$

In what follows we assume that the empirical constraints $\vec{a}$ and $\bar{p}(y)$ correspond to their true values, i.e. $p(x,y) \in \mathcal{P}(\vec{a})$. While this cannot be exactly true, the estimated expected values converge to the true ones (see [15]), and this deviation can be controlled via standard statistical methods. Since $p(x,y) \in \mathcal{P}(\vec{a})$ its information must be larger than that of $p_{MI}(x,y)$, and thus

$$e^*(p) \leq \frac{1}{2}\left(H(Y) - I[p_{MI}(x,y)]\right) . \qquad (15)$$

We thus have a model free bound on the Bayes error of the unknown distribution $p(x,y)$. An obvious shortcoming of the above bound is that it does not relate to the classification error under when using the *plug in* distribution $p_{MI}(y|x)$ as the class predictor. Denote this error by $e_{MI}$. Then using Equation 13 with $q(y|x) = p_{MI}(y|x)$ we have

$$e_{MI} \leq -\sum p(x,y)\log_2 p_{MI}(y|x) . \qquad (16)$$

But the special form of $p_{MI}(y|x)$ implies that we can replace expectation over $p(x,y)$ with expectation over $p_{MI}(x,y)$, when $p_{MI}(x)$ is strictly positive

$$\sum p(x,y)\log_2 p_{MI}(y|x) = -\sum p_{MI}(x,y)\log_2 p_{MI}(y|x) .$$

The RHS is the conditional entropy $H[p_{MI}(y|x)]$, which implies the following bound

**Proposition 2** *If $p_{MI}(x) > 0$ for all $x$, the generalization error of the classifier based on $p_{MI}(y|x)$ satisfies*

$$e_{MI} \leq H(Y) - I[p_{MI}(x,y)] . \qquad (17)$$

Note that the bound on the optimal Bayes error of the true distribution is tighter than the above bound by a factor of 2. It will be interesting to see whether these bounds can be improved.

## 6 Relation to Other Methods

As seen previously, MinMI provides a model for $p_{MI}(y|x)$ that is similar to the one obtained in both conditional and generative modeling. We now expand on the differences between these methods.

### 6.1 Maximum Entropy of the Joint Distribution

The joint entropy of $X$ and $Y$ is related to the mutual information via

$$I(X;Y) = H(X) + H(Y) - H(X,Y) . \qquad (18)$$

Thus, if both marginals are assumed to be known, the problems of Maximum Entropy and Minimum Mutual Information coincide. The model of the joint distribution in this case has the following form

$$p_{ME}(x,y) = \frac{1}{Z}e^{\vec{\phi}(x)\vec{\psi}(y)+A(x)+B(y)} . \qquad (19)$$

where $A(x)$ is a free parameter which is adjusted so that $p_{ME}(x,y)$ has the desired marginal over $X$.

The resulting conditional model is then

$$p_{ME}(y|x) = \frac{1}{Z_x}e^{\vec{\phi}(x)\vec{\psi}(y)+A(x)+B(y)} . \qquad (20)$$

When the marginal over $X$ is not known, but $p(y)$ is, maximizing the joint entropy is equivalent to maximizing $H(X|Y)$, which is equivalent to maximizing $H(X|Y=y)$ for each value of $y$ independently. Note that under this approach, changing the values of $\vec{a}(y)$ for a given value of $y$ will note change $p(X|y)$ for other values of $y$. This does not seem to be a desirable property, and does not hold in the MinMI case. One example of maximizing joint entropy is the Naive Bayes model which results from maximizing $H(X|Y)$ subject to a constraint on conditional singleton marginals, and the class marginals.

### 6.2 Conditional Random Fields and Logistic Regression

Conditional Random Fields (CRF) are models of the conditional distribution

$$p_\lambda(y|x) = \frac{1}{Z_\lambda(x)}e^{\sum_{k=1}^d \lambda_k f_k(x,y)} . \qquad (21)$$

The $d$ functions $f_k(x,y)$ are assumed to be known in advance, or are chosen from some large set. This becomes similar to our setting if one chooses functions

$$f_{i,y_j}(x,y) = \delta_{y,y_j}\phi_i(x) . \qquad (22)$$

In fact, the MinMI formalism could be equally applied to general functions of $X$ and $Y$ as in CRFs. We focus on functions of $X$ for ease of presentation.

CRFs are commonly trained using by choosing $\lambda_i$ which maximize the conditional maximum likelihood [13] given by

$$\sum_{x,y} \bar{p}(x,y) \log p_\lambda(y|x) = -\langle \log Z_\lambda(x) \rangle_{\bar{p}(x)} + \sum_{k=1}^{d} \lambda_k \langle f_k \rangle_{\bar{p}(x,y)} ,$$

where $\bar{p}(x,y)$ is the empirical distribution.

This target function is seen to depend on the empirical expected values of $f_k$ but also on the empirical marginal $\bar{p}(x)$. This is of course true for all conditional logistic regression models, and differentiates them from MinMI, which has access only to the expected values of $\vec{\phi}(x)$.

### 6.3 Constraints on Marginals

Models of distributions over large sets of variables often focus on the marginal properties of subsets of these variables. Furthermore, maximum likelihood estimation over Markov fields is known to be equivalent to matching the empirical marginals of the cliques in the graph. We now define the MinMI version of the marginal matching problem.

Denote by $X \equiv (X_1, \ldots, X_n)$ an $n$ dimensional feature vector, and by $\{X_C\}$ a set of subsets of variables of $X$ (e.g. all singletons or pairs of $X_i$). Assume we are given the empirical conditional marginals $p(X_C|Y)$. In our notation, this is equivalent to choosing the following $\vec{\phi}$

$$\phi_{x_C}(\hat{x}) = \delta_{\hat{x}_C, x_C} . \qquad (23)$$

which has the expected value $p(x_C|y)$.

The MinMI distribution in this case would have the following form

$$p_{MI}(y|x) = \bar{p}(y) e^{\sum_{x_C} \psi(x_C,y) + \gamma(y)} . \qquad (24)$$

## 7 MinMI Algorithms

In order to find the classification distribution $p_{MI}(y|x)$ the optimization problem in Equation 5 or its dual in Equation 10 need to be solved. This section describes several algorithmic approaches to calculating $p_{MI}(y|x)$. When $|X|$ is small enough to allow $O(|X|)$ operations, exact algorithms can be used. Otherwise, random sampling techniques are used to overcome complexity issues.

### 7.1 Solving the Primal Problem

The characterization of $p_{MI}(x|y)$ is similar to that of the Rate Distortion channel [2] or the related Information Bottleneck distribution in [17]. There are iterative procedures for finding the optimal distributions in these cases, although usually as a function of the Lagrange multipliers (i.e. $\vec{\psi}(y)$) rather than of the value of the constraints. In what follows we outline an algorithm which finds $p_{MI}(x|y)$ for any set of empirical constraints.

The basic building block of the algorithm is the *I-projection* [3]. The *I-projection* of a distribution $q(x)$ on a set of distributions $\mathcal{F}$ is defined as the distribution $p^* \in \mathcal{F}$ which minimizes the KL-divergence to the distribution $q(x)$ : $p^* \equiv \arg\min_{p \in \mathcal{F}} D_{KL}[p|q]$. When $\mathcal{F}$ is determined by expectation constraints

$$\mathcal{F}(\vec{\phi}(x), \vec{a}) \equiv \left\{ p(x) : \langle \vec{\phi}(x) \rangle_{p(x)} = \vec{a} \right\}$$

the projection is given by

$$p^*(x) = \frac{1}{Z_\lambda^*} q(x) e^{\vec{\lambda}^* \cdot \vec{\phi}(x)} , \qquad (25)$$

where $\vec{\lambda}^*$ are a set of Lagrange multipliers, chosen to fit the desired expected values, and $Z_\lambda^*$ is a normalization factor. The values of $\vec{\lambda}^*$ can be found using several optimization techniques such as Generalized Iterative Scaling [4] or gradient based methods. All projection algorithms involve the computation of the expected value of $\vec{\phi}(x)$ under distributions of the form $q(x) e^{\vec{\lambda} \cdot \vec{\phi}(x)}$.

The similarity between the form of the projection in Equation 25 and the characterization of $p_{MI}(x|y)$ in Equation 6, implies that $p_{MI}(x|y)$ is an *I-projection* of $p_{MI}(x)$ on the set $\mathcal{F}(\vec{\phi}(x), \vec{a}(y))$. The fact that $p_{MI}(x)$ is dependent on $p_{MI}(x|y)$ through marginalization implies an iterative algorithm where marginalization and projection are performed. This procedure, is described in Figure 1. It can be shown to converge using the Pythagorean property of the *I-projection* as in [6].

The above algorithm cannot be implemented in a straightforward manner when $|X|$ is large, since it involves an explicit representation of $p_t(x)$. To circumvent this problem, we note that applying the primal algorithm recursively results in the following representation of $p_t(x)$ as a mixture of $|Y|^t$ elements

$$p_t(x) = \sum_{\vec{y} = (y_1, \ldots, y_t)} c(\vec{y}) p_{\vec{y}}(x) , \qquad (27)$$

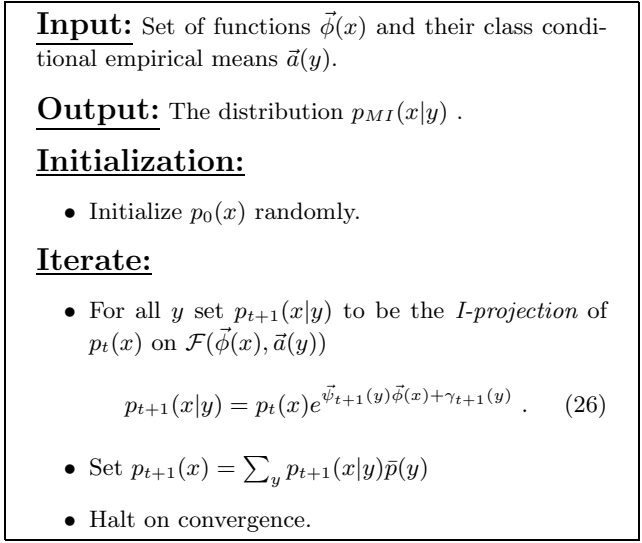

**Input:** Set of functions $\vec{\phi}(x)$ and their class conditional empirical means $\vec{a}(y)$.

**Output:** The distribution $p_{MI}(x|y)$ .

### Initialization:
- Initialize $p_0(x)$ randomly.

### Iterate:
- For all $y$ set $p_{t+1}(x|y)$ to be the *I-projection* of $p_t(x)$ on $\mathcal{F}(\vec{\phi}(x), \vec{a}(y))$

$$p_{t+1}(x|y) = p_t(x)e^{\vec{\psi}_{t+1}(y)\vec{\phi}(x) + \gamma_{t+1}(y)} \ . \quad (26)$$

- Set $p_{t+1}(x) = \sum_y p_{t+1}(x|y)\bar{p}(y)$
- Halt on convergence.

Figure 1: An algorithm for solving the primal problem.

where

$$\begin{aligned} c(\vec{y}) &\equiv Z(\vec{y}) \prod_{\hat{t}=1:t} \bar{p}(y_{\hat{t}}) e^{\gamma_{\hat{t}}(y_{\hat{t}})} & (28) \\ p_{\vec{y}}(x) &\equiv \frac{1}{Z(\vec{y})} e^{\vec{\Psi}(\vec{y}) \cdot \vec{\phi}(x)} \ , \quad \vec{\Psi}(\vec{y}) \equiv \sum_{\hat{t}=1:t} \vec{\psi}_{\hat{t}}(y_{\hat{t}}) \end{aligned}$$

and $Z(\vec{y})$ is the partition function normalizing $e^{\vec{\Psi}(\vec{y}) \cdot \vec{\phi}(x)}$.

We are still left with a number of parameters exponential in $T$. To overcome this difficulty we use random sampling to draw elements of the mixture $p_t(x)$. Such random sampling can be performed using the fact that $c(\vec{y})$ is a distribution over the vector $\vec{y}$ and thus we can use any sampling technique (here we used Gibbs) for $\vec{y}$ to draw elements in the mixture according to $c(\vec{y})$. After drawing $N$ elements from $c(\vec{y})$ we approximate $p_t(x)$ using

$$p_t(x) \approx \frac{1}{N} \sum_{n=1}^{N} \frac{1}{Z(\vec{y}^n)} e^{\vec{\Psi}(\vec{y}^n) \cdot \vec{\phi}(x)} \ . \quad (29)$$

Finally, performing the *I-projection* of the estimated $p_t(x)$ requires calculating the expected value of $\vec{\phi}(x)$ under distributions of the form $p_{\vec{y}}(x) e^{\vec{\phi}(x) \cdot \vec{\lambda}(y)}$. This can often be done without explicitly summing over $X$. For example, when $\vec{\phi}(x)$ represent singleton marginal constraints (see Section 6.3) the expected values of $\vec{\phi}(x)$ correspond to marginals over simple Markov Fields, and these can be easily calculated.

One shortcoming of the above algorithm is that it requires storage of all Lagrange multipliers calculated in previous iterations. It could thus become costly as the number of iterations grow. However, in our experimental evaluations we found that under 50 iterations are sufficient for the algorithm to converge. The dual algorithm, presented next, does not have this dependence on the number of iterations.

In [12] it was shown how a related problem can be solved with iterative MCMC without storage. It will be interesting to see whether this approach can be applied here.

### 7.2 Solving the Dual Problem

The dual problem as given in Equation 10 is a geometric program and as such can be solved efficiently using interior point algorithms [1]. When $|X|$ is too large to allow $O(|X|)$ operations, such algorithms are no longer practical. However, oracle based algorithms such as the Ellipsoid algorithm or Cutting Plane Methods [9] are still applicable (in our experiments we used the ACCPM package described in [9]). The above algorithms require an oracle which specifies if a given point is feasible, and if not, specifies a constraint which it violates. For the constraints in Equation 10 this amounts to finding the $x$ maximizing the constrained function

$$\begin{aligned} x_{max} &\equiv \arg \max_x f(x) \\ f(x) &\equiv \sum_y \bar{p}(y) e^{\gamma(y) + \vec{\psi}(y) \cdot \vec{\phi}(x)} \ . \end{aligned}$$

The point $(\gamma(y), \vec{\psi}(y))$ is then feasible if $x_{max} \leq 1$. Since $f(x)$ may be interpreted as an unnormalized distribution over $x$, finding $x_{max}$ is equivalent to finding its maximum probability assignment. This is known as the MAP problem in the AI literature, and can be tackled using random sampling techniques as in [16] [3].

The primal and dual algorithms gave similar results, although for the experiments described here the primal algorithm converged faster.

## 8 Illustrative example

To demonstrate some properties of the MinMI solution, we apply it to the well known problem of constraints on the first and second moments of a distribution. The MaxEnt solution to the above problem would be a Gaussian model of $p(x|y)$ with the appropriate mean and variance. The MinMI solution to this problem is shown in Figure 2, and is quite different from a Gaussian [4].

---
[3]We used the ML assignment as an initial guess, followed by Gibbs sampling

[4]The exact solution should be two delta functions, but due to numerical precision issues the algorithm converges to the distribution shown here.

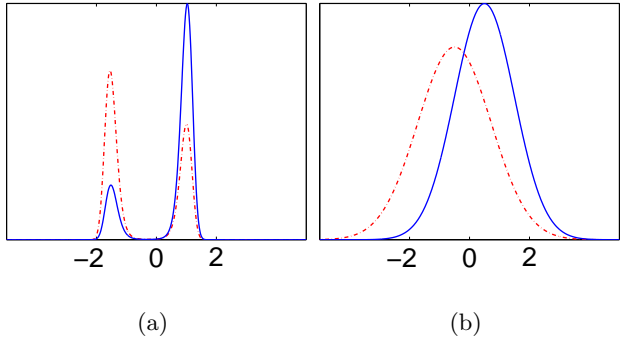

Figure 2: Application of MinMI to the first and second moment problem. Here $\vec{\phi}(x) = [x, x^2]$, $\vec{a}(y_1) = [-0.5, 1.75]$, $\vec{a}(y_2) = [0.5, 1.25]$ and $\bar{p}(y)$ is uniform. The $x$ range is 500 equally spaced points between $-5$ and $5$. **a.** The conditional distributions $p_{MI}(x|y_1)$ (red dashed line), $p_{MI}(x|y_2)$ (blue solid line). **b.** The MaxEnt solution for the given problem.

The two distributions $p_{MI}(x|y_1), p_{MI}(x|y_2)$ are structured to obey the moment constraints imposed by $\vec{a}(y)$ while keeping as little information as possible about the identity of $y$. This is done by concentrating most of their joint mass around two points.

## 9  Experimental Results

We tested the MinMI classification scheme on 12 datasets from the UCI repository [14]. Only the discrete features in each database were considered. The algorithm for the primal problem was used (see Section 7.1) with $N = 5000$ Gibbs sampling draws on each iteration. The features used as input to the MinMI algorithm were the singleton marginal distributions of each of the features, as described in Section 6.3. Classification performance was compared to that of Naive Bayes [5] and the corresponding first order conditional Log-Linear model [6]. Note that these two models use a parametric form of $p(y|x)$ that is nearly identical to that of MinMI (they differ only in that they have a partition function $Z_x$).

The results for all the datasets are shown in Figure 3. It can be seen that except for one database (heart-disease) MinMI performs similar to, and usually better than Naive Bayes. Also, both methods outperform the Log-Linear model on small sample sizes as described previously in [15] (Log-Linear outperforms MinMI and Naive Bayes only on 3/12 databases for small sample sizes). Table 1 shows the generalization error measured using 10 fold cross validation. It is not surprising that the Log-Linear model outperforms both Naive Bayes and MinMI, since its theoretical asymptotic error is lower than theirs, and the sample size is large enough for it to generalize well. However, MinMI achieves $98 \pm 2\%$ of the Log-Linear model performance, compared to $94 \pm 7\%$ for Naive Bayes.

---

[5]Marginals used for Naive Bayes and MinMI were estimated using Laplace smoothing with a pseudo-count of 1.

[6]In the linearly separable case the conditional model solution is not unique. As in [15] we randomly sample separating hyperplanes, by carrying out a random walk in version space. The reported performance is the average generalization error over the sampled hyperplanes.

| Name | MinMI | Naive Bayes | Log-Linear |
|---|---|---|---|
| voting-records | $5.06 \pm 0.01$ | $10.65 \pm 0.18$ | $4.26 \pm 0.32$ |
| breast-cancer | $27.78 \pm 0.92$ | $28.88 \pm 0.44$ | $27.24 \pm 0.89$ |
| sick | $6.11 \pm 0.00$ | $6.11 \pm 0.00$ | $6.11 \pm 0.00$ |
| splice | $7.40 \pm 0.48$ | $21.84 \pm 0.15$ | $4.08 \pm 0.11$ |
| kr-vs-kp | $6.06 \pm 0.31$ | $22.53 \pm 0.16$ | $5.60 \pm 0.66$ |
| promoters | $9.30 \pm 2.02$ | $22.35 \pm 1.44$ | $8.28 \pm 1.56$ |
| hepatitis | $19.28 \pm 0.84$ | $19.56 \pm 0.58$ | $17.46 \pm 0.86$ |
| heart-disease | $24.91 \pm 1.41$ | $18.13 \pm 0.38$ | $19.17 \pm 0.88$ |
| credit | $14.49 \pm 0.01$ | $14.46 \pm 0.22$ | $13.80 \pm 0.39$ |
| adult | $20.48 \pm 0.09$ | $22.38 \pm 0.04$ | $17.34 \pm 0.02$ |
| lymphography | $18.59 \pm 1.20$ | $18.56 \pm 0.81$ | $16.34 \pm 1.56$ |
| hypo | $7.73 \pm 0.02$ | $7.71 \pm 0.00$ | $7.71 \pm 0.00$ |

Table 1: Results (percent error) of 10 fold cross validation experiments on the UCI datasets. Confidence intervals are standard deviations over 40 different cross validation partitions. MinMI significantly outperforms Naive Bayes on 5/12 datasets. Naive Bayes significantly outperforms MinMI only on the heart-disease dataset.

MinMI may outperform Log-Linear models on datasets with large number of features, where the latter are more likely to overfit.

## 10  Discussion

We introduced the principle of minimum mutual information (MinMI) as a fundamental method for inferring a joint distribution in the presence of empirical conditional expectations. This principle replaces Maximum Entropy for such cases and in general is not equivalent to a maximum likelihood estimation of any parametric model.

It is interesting to note that the MinMI solution for a multivariate $X$ does not satisfy the conditional independence properties which the corresponding graphical model possesses. This is clear already when singleton marginals are used as constraints. The resulting $p_{MI}(x|y)$ may in fact contain elaborate dependencies between the variables. To see why this comes about consider the extreme case where all the conditional singleton marginals are constrained to be equal. It is easy to see that under $p_{MI}(x|y)$ the variables $X_1, \ldots, X_n$ will be completely dependent (i.e. $p_{MI}(x_1, \ldots, x_n|y) = p_{MI}(x_1|y)$).

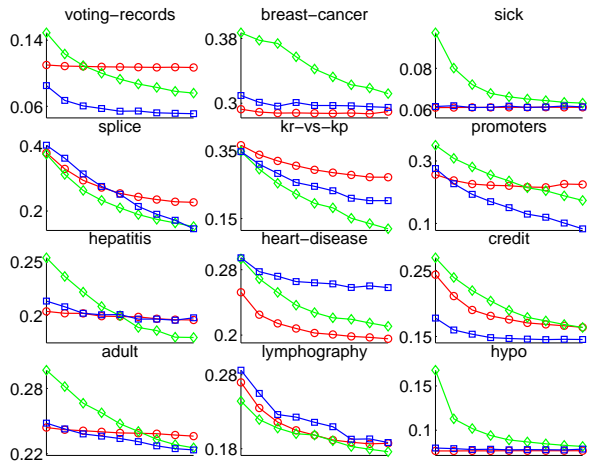

Figure 3: Classification error as a function of the number of samples used in training, for several UCI datasets. MinMI is blue line and squares, Naive Bayes is red line and circles, Log-Linear model is green line and diamonds. For each sample size, 1000 random splits of the data were performed. The samples sizes in the plots are $20, 30, \ldots, 100$.

It is important to stress that $p_{MI}(x)$ is not argued to be a *model* of the true underlying distribution. Rather, as the game theoretic analysis shows, it represents a worst case scenario with respect to prediction.

Although we did not address the case of continuous $X$ domain directly, our formalism applies there as well. Consider a vector of continuous variables, with constraints on the means and covariances of subsets of its variables. The MinMI distribution in this case will be related to the corresponding Gaussian Markov field.

Another natural extension of the current work is feature induction [5]. As was done in [6], one can look for features $\vec{\phi}(x)$ which maximize the minimum mutual information calculated in the current work. In [6] both marginals were assumed to be known. The extension to unknown marginals should provide a powerful tool for feature induction over large variables sets.

## Acknowledgements

We are grateful to the anonymous reviewer who suggested the game theoretic interpretation based on [7]. We thank R. Gilad-Bachrach for helpful discussions, and G. Chechik and A. Navot for reading the manuscript. This work is partially supported by a grant from the Israeli Academy of Science (ISF). We wish to thank the University of Pennsylvania for its hospitality during the writing of this paper.

## References

[1] M. Chiang and S. Boyd. Geometric programming duals of channel capacity and rate distortion. *IEEE Trans. on Information Theory*, 50(2):245–258, 2004.

[2] T.M. Cover and J.A Thomas. *Elements of information theory*. Wiley, 1991.

[3] I. Csiszar. I-divergence geometry of probability distributions and minimization problems. *Annals of Probability*, 3(1):146–158, 1975.

[4] J.N. Darroch and D. Ratcliff. Generalized iterative scaling for log-linear models. *Ann. Math. Statist.*, 43:1470–1480, 1972.

[5] S.A. Della-Pietra, V.J. Della-Pietra, and J.D. Lafferty. Inducing features of random fields. *IEEE Transactions on PAMI*, 19(4):380–393, 1997.

[6] A. Globerson and N. Tishby. Sufficient dimensionality reduction. *Journal of Machine Learning Research*, 3:1307–1331, 2003.

[7] D. Grnwald and A.P. Dawid. Game theory, maximum entropy, minimum discrepancy, and robust bayesian decision theory. *Annals of Statistics*, To appear, 2004.

[8] M.E. Hellman and J. Raviv. Probability of error, equivocation, and the chernoff bound. *IEEE Transactions on Information Theory*, 16(4):368–372, 1970.

[9] J. Gondzio J., O. du Merle, R. Sarkissian, and J.P. Vial. Accpm - a library for convex optimization based on an analytic center cutting plane method. *European Journal of Operational Research*, 94:206–211, 1996.

[10] T Jaakkola, M. Meila, and T. Jebara. Maximum entropy discrimination. In *Advances in Neural Information Processing Systems 12*, 1999.

[11] E.T. Jaynes. Information theory and statistical mechanics. *Physical Review*, 106:620–630, 1957.

[12] L.A. Wasserman J.D. Lafferty. Iterative markov chain monte carlo computation of reference priors and minimax risk. In *Proc. of the 17th Conference on Uncertainty in Artificial Intelligence*, 2001.

[13] J. Lafferty, A. McCallum, and F. Pereira. Conditional random fields: Probabilistic models for segmenting and labeling sequence data. In *Proc. 18th ICML*, pages 282–289, 2001.

[14] P. Murphy and D. Aha. Uci repository of machine learning databases (machine readable data repository. *UCI Dept. of Info. and Comp. Sci.*

[15] A.Y. Ng and M.I. Jordan. On discriminative vs. generative classifiers. In *Advances in Neural Information Processing Systems 14*, pages 605–610, 2001.

[16] J. Park and A. Darwiche. Approximating map using stochastic local search. In *Proc. of the 17th Conference on Uncertainty in Artificial Intelligence*, 2001.

[17] N. Tishby, F.C. Pereira, and W. Bialek. The information bottleneck method. In *Proc. of the 37-th Annual Allerton Conference on Communication, Control and Computing*, pages 368–377, 1999.